\documentclass[letterpaper, 10 pt, conference]{ieeeconf}  

\IEEEoverridecommandlockouts  
\usepackage[pdftex]{graphicx}
\DeclareGraphicsExtensions{.pdf,.jpeg,.png}
\usepackage{amsmath} 
\usepackage{amssymb}  

\usepackage{tabularx,booktabs,multirow}
\usepackage{array}
\usepackage{float}
\usepackage{makecell}
\usepackage{tablefootnote}
\usepackage{physics}
\usepackage{multicol}
\usepackage{booktabs}
\usepackage{threeparttable}
\usepackage{url}
\usepackage[font=small,skip=10pt]{caption}
\usepackage[dvipsnames]{xcolor}
\usepackage{amsmath}
\usepackage{xcolor}
\usepackage{algorithmic}
\usepackage{dirtytalk}
\usepackage[ruled,vlined]{algorithm2e}

\newcolumntype{P}[1]{>{\centering\arraybackslash}p{#1}}
\SetKwInOut{Input}{input}
\SetKwInOut{Output}{output}
\makeatletter
\let\NAT@parse\undefined
\makeatother
\usepackage{hyperref}
\hypersetup{colorlinks,allcolors=red}

\title{\LARGE \bf
Collaborating for Success: Optimizing System Efficiency and Resilience Under Agile Industrial Settings
}

\author{Sunny Katyara$^{1}$, Suchita Sharma$^{2}$,  Praveen Damacharla$^{3}$, Carlos Garcia$^{1}$, Francis O'Farrell$^{1}$, Philip Long$^{1}$  
\thanks{This research is supported by EU funded 'ACROBA' and 'CORESENSE' projects under grant agreement No 101017284 and 101070254 respectively. \textit{(Corresponding author: Sunny Katyara)} }
\thanks{$^{1}$ Department of Robotics \& Automation, Irish Manufacturing Research Company Ltd, Ireland (e-mail: {\tt\small name.surname@imr.ie
}).}
\thanks{$^{2}$ Software Engineering Division, Shutterfly, USA (e-mail: {\tt\small suchita86@gmail.com
}).}
\thanks{$^{3}$ KineticAI INC., Woodlands Texas, USA ({e-mail: \tt\small praveen@kineticai.com}).}
}

\begin{document}

\maketitle
\thispagestyle{empty}
\pagestyle{empty}

\begin{abstract}

Designing an efficient and resilient human-robot collaboration strategy that not only upholds the safety and ergonomics of shared workspace but also enhances the performance and agility of collaborative setup presents significant challenges concerning environment perception and robot control. In this research, we introduce a novel approach for collaborative environment monitoring and robot motion regulation to address this multifaceted problem. Our study proposes novel computation and division of safety monitoring zones, adhering to ISO 13855 and TS 15066 standards, utilizing 2D lasers information. These zones are not only configured in the standard three-layer arrangement but are also expanded into two adjacent quadrants, thereby enhancing system uptime and preventing unnecessary deadlocks. Moreover, we also leverage 3D visual information to track dynamic human articulations and extended intrusions. Drawing upon the fused sensory data from 2D and 3D perceptual spaces, our proposed hierarchical controller stably regulates robot velocity, validated using Lasalle in-variance principle. Empirical evaluations demonstrate that our approach significantly reduces task execution time and system response delay, resulting in improved efficiency and resilience within collaborative settings.  

\end{abstract}

\section{INTRODUCTION}
\textbf{Human-robot collaboration (HRC)} represents a paradigm that capitalizes on the complementary abilities of human cognition and fine-motor skills, combined with the repeatability, strength, and speed of robotic machines, to achieve complex tasks with enhanced efficiency and effectiveness. Collaborative robots, commonly known as cobots, have assumed a critical role in the agile manufacturing industry, fostering improved competitiveness and resilience in the face of mass customization and market volatility. The widespread adoption of cobots can be attributed to their accessibility, programmability, and reconfigurability, leading to a projected application value of $\$ 3.2$ billion by the end of 2025 \cite{b1}. The International Federation of Robotics (IFR) report for 2023 indicates a remarkable 50\% growth in the installation of cobots across various industrial applications, with 39,000 out of over 478,000 installations constituting cobots \cite{b2}. This significant growth trajectory is expected to persist, with a projected 88\% increase in cobot installations by 2025 \cite{b3}. Such a trend is closely aligned with the emergence of Industry 5.0, a manufacturing paradigm that places emphasis on the deployment of human-centric, flexible, and resilient machines like cobots.

   \begin{figure}[t]
      \centering
      \includegraphics[width=8.5cm]{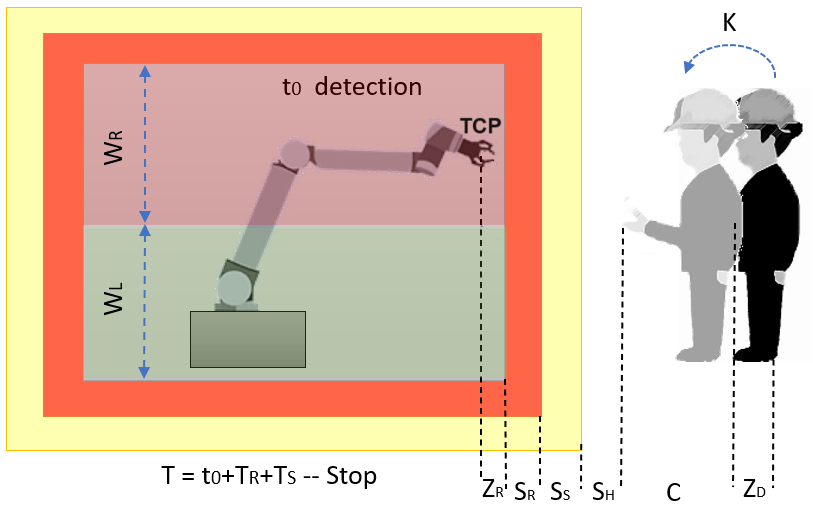}
      \caption{Proposed configuration and parameterization of safety zones for SSM mode in accordance with ISO 13855 and TS 15066.}
      \label{fig2}
   \end{figure} 

HRC presents several challenges, with two critical being (i) the proper placement and configuration of workspace monitoring devices, and (ii) the design and integration of robust interaction controllers for agile behavior. The assurance of stable and safe interaction is of utmost importance to achieve higher precision, repeatability, and productivity gains, while simultaneously ensuring the dependability and trustworthiness of the shared workspace. In addressing the challenge of safe interaction, monitoring devices must adhere to ISO 13851 and be designed to safeguard human co-workers from injury, all while maintaining optimal workspace ergonomics, as depicted in Fig. \ref{fig2} (the proposed setup). Whereas, for stable interaction, the controller must exhibit the global asymptotic equilibrium, validated using LaSalle's invariance principle \cite{b4}. The choice of industrial-graded monitoring device for safeguarding i.e., light curtains and laser scanners, hinges on the specific nature of the task and the severity of the operation. For applications necessitating high-resolution monitoring in proximity to hazards and immediate tripping of the robot, light curtains prove suitable. Conversely, laser scanners find typical use in segmented operations to curtail frequent and unnecessary downtime of cobots. Cobots are designed to work in four collaborative modes including Safety-rated Monitored Stop (SMS), Hand Guiding (HG), Speed and Separation Monitoring (SSM), and Power and Force Limiting (PFL), considering their speed, reach, and payload limits, as per ISO 10218-1/2 standard, ensuring safe interaction with humans \cite{b6}.

\section{Related Works and Contributions}

In \cite{c1}, a human-aware manipulation planner is introduced to promote safe and socially acceptable interactions in human-centric settings. Using 2D lasers and 3D depth sensors, it generates collision-free trajectories for cobots during parts handover tasks. Yet, its focus on position-level control, instead of kinematic-dynamic planning according to ISO 13855, led to increased cycle times and deadlocks. In previous work \cite{c1b}, we introduce security system for industrial environments, where cobots uses 2D lasers for spatial monitoring. Despite its efficiency, it overlooks human interactions in 3D spaces and lacks adaptability under stochastic conditions. Contrastingly, \cite{c2} proposes a method using sensor-less collision detection for an industrial-grade polishing task. It facilitates ergonomic workspace sharing, but its agility and performance are not compared against human-human cooperation. Our another work in \cite{c3} presents a hierarchical velocity controller for cobots, employing visual-tactile fusion for task management. While it improves ergonomics, it lacks in system redundancy, resilience and human-centric optimization. In \cite{c4}, a method combines human-tracking and tactile sensing, utilizing deep neural networks, for safe human-cobot collaboration. Despite improving safety and agility, it does not enhance task performance or system robustness. In \cite{c5}, an agent-based framework dynamically allocates tasks between humans and robots, showcasing robustness against uncertainties. However, it does not optimize task performance in shared workspace. Whereas, \cite{c6} allows industrial robots to operate collaboratively under the SSM mode. Using SafetyEye (i.e., safe vision system) for flexible zone monitoring, it dynamically adapts to human movements and expedited task execution, but yet system resilience. flexibility and human-centricity are not co-optimized.

To the best of our knowledge, no existing framework ensures the concurrent achievement of safety, ergonomics, task execution performance, and system effectiveness in human-robot collaboration within agile manufacturing settings. This research proposes a holistic framework that addresses these challenges by implementing the modified SSM collaborative mode, utilizing a combination of 2D laser scanners and 3D depth sensors for collaborative part sorting and product assembly. To achieve this, the laser scanners are optimally positioned at strategic locations to cover maximum of the shared workspace (i.e., 400 mm above the ground and 100 mm with respect to robot base frame according to EN 61496), while the depth sensors are employed to robustly localize candidate objects and track human skeleton, irrespective of their shape, height, posture, and walking speed. The study contributes to the field of HRC in two ways:

\begin{itemize}
    \item A novel configuration of safety zones is proposed, which not only complies with ISO 13855 and TS 15066 but also enhances the system's up-time and productivity. The proposition is experimentally evaluated and empirically validated through quantitative KPIs, i.e., cycle time (CT) and overall equipment effectiveness (OEE).
    \item A blended hierarchical velocity controller is formulated to modulate cobot's motion based on human movements within collaborative workspace, taking into account both the 2D laser and 3D depth information. The robustness of task execution and safety alignment are assessed using metrics i.e., flexibility rate (FR), and reaction time (RT). In addition, its asymptotic stability, assessed using Lasalle theorem, is reported for a collaborative scenario.  
\end{itemize}

\section{Scientific Methodology}

\subsection{Safety Zones Configuration}

   \begin{figure}[t]
      \centering
      \includegraphics[width=8.5cm]{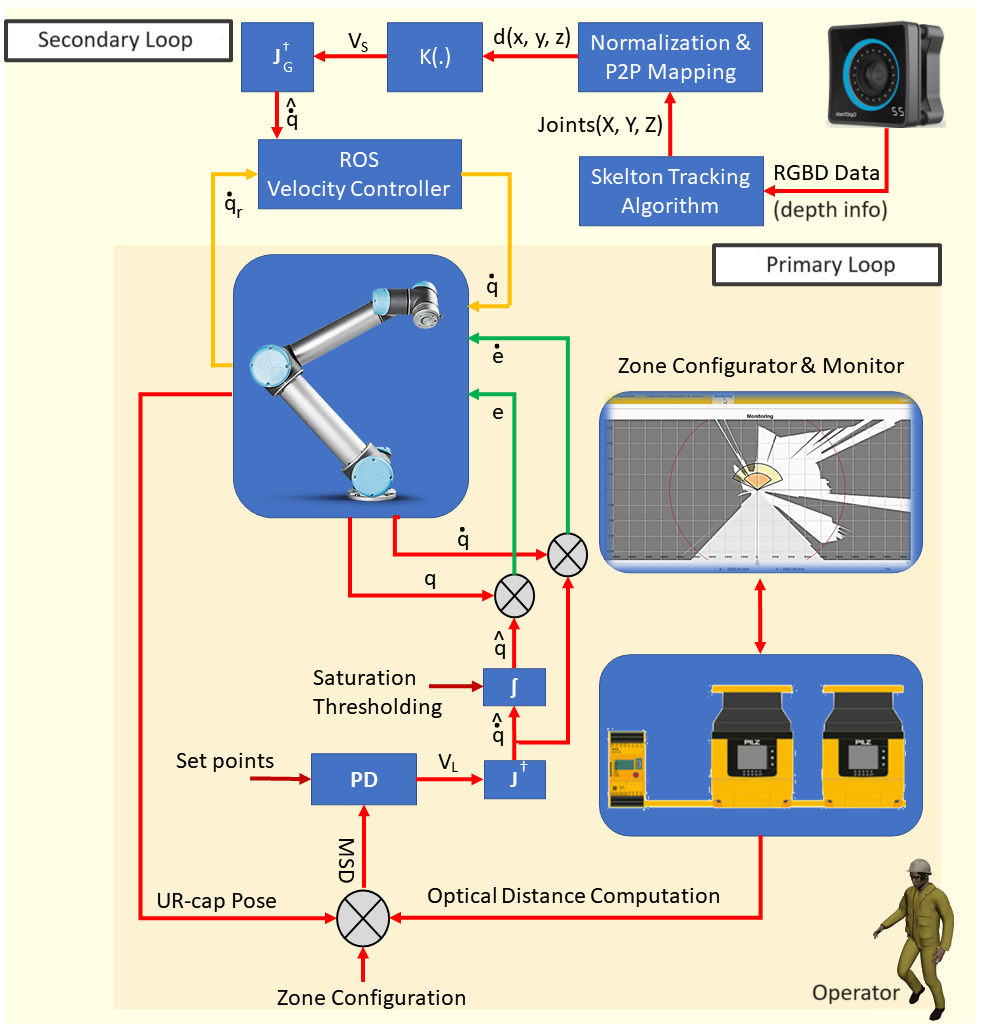}
      \caption{Hierarchical velocity control architecture for fused perceptual monitoring and collaborative decision making.}
      \label{fig3}
   \end{figure}

   \begin{figure*}[t]
      \centering
      \includegraphics[width=18cm]{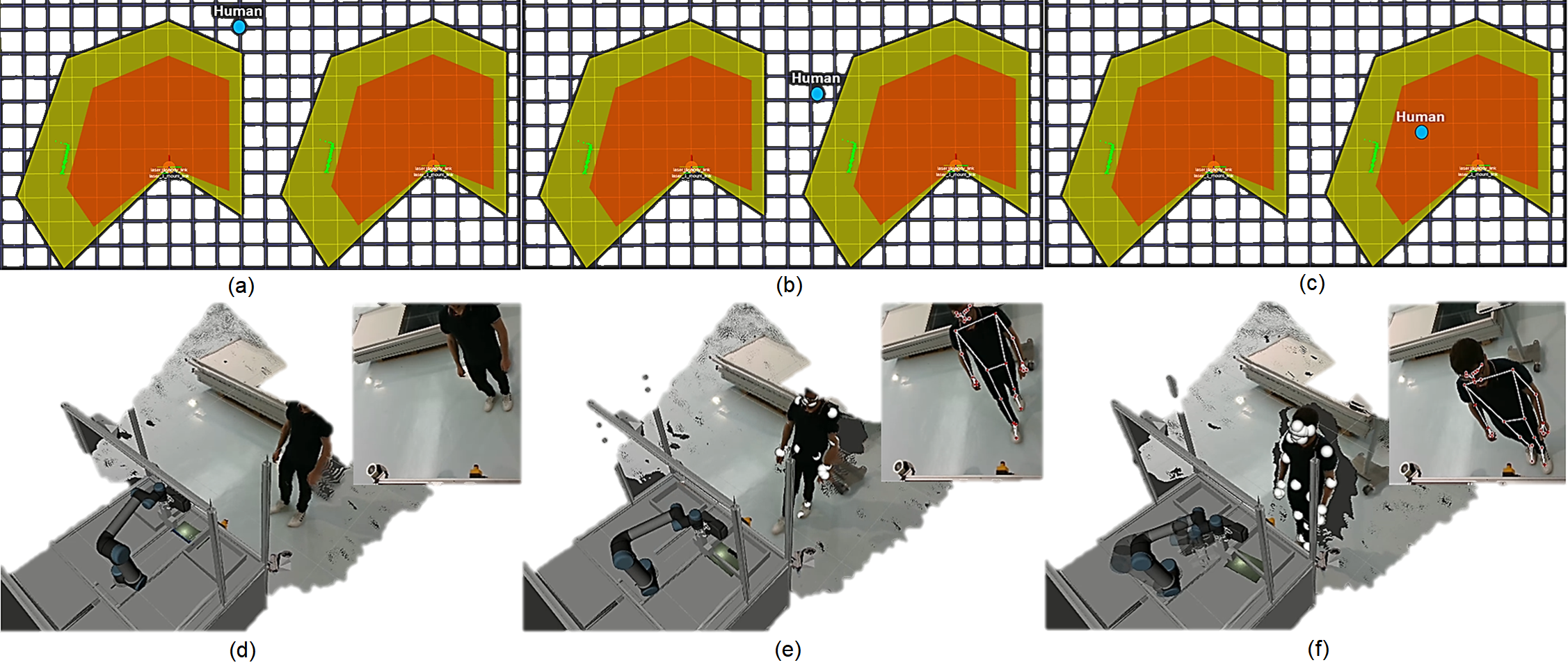}
      \caption{Collaborative workspace monitoring using 2D laser scanners and 3D depth sensor. (a-c) demonstrate configuration of safety zones for human activity tracking using strategically positioned scanners (white is normal, yellow is warning and red is dangerous), and (d-f) illustrate precise tracking of operator's skeleton to facilitate improved cooperation and resilience.}
      \label{figK}
   \end{figure*}

To facilitate the implementation of modified SSM collaborative mode, accurately determining the MSD (minimum separation distance) between the cobot and human co-worker is imperative to mitigate the risk of collisions during operation. Failure to comply with specified MSD will lead to cessation of operation. The calculation of the MSD is determined by applying the formula described in (\ref{eq1}), which takes into consideration various parameters, including the level of human activeness, the speed of cobot, its payload, its reachability, and the time taken by system to stop. These parameters are outlined in accordance with guidelines specified in ISO 13855.  

\begin{equation}
    MSD=(K{\times}T)+C+\delta
    \label{eq1}
\end{equation}

where K represents the speed at which the operator approaches the cobot, T denotes the overall response time of the perception and control system to bring the cobot to a complete stop. Additionally, C represents the intrusion distance of articulated body parts of the operator reaching the danger zone, i.e., the height of lower limbs and length of upper limbs. The symbol $\delta$ denotes the allowance for position uncertainty of both the operator and the cobot, arising from exteroceptive and proprioceptive inaccuracies. 

In this context, the MSD is defined as distance between the TCP (tool center point) of the cobot's extended arm and the inner edge of the yellow zone, as illustrated in Fig. \ref{fig2}. However, the height, along with the length and width of the safety zones, constitutes another crucial factor that must be taken into account. The height consideration should account for potential intrusion of human bodies, the flooring of cobot, and its working manifold. While ISO 13855 considers the working manifold of the cobot as a whole subspace, it yields sub-optimal results under complex industrial scenarios, such as discrete and sequential assembly, progressive part inspection, and object co-carrying, among others due to supererogatory safety configurations and dynamic human articulations. To maximize cobot's utilization and achieve optimal operation, it is necessary to segment working manifold into quadrants. In this research, we propose the segmentation of the working manifold into two equal quadrants (i.e., 425 mm), corresponding to half of the kinematic reach of cobot (UR5), as depicted in Fig. \ref{fig2}. This segmentation allows for greater flexibility in the cobot's operation.

The established zone settings are static and optimally implemented using 2D scanners. However, for 3D workspace monitoring, a dynamic approach to computing run-time safety distances becomes necessary, owing to extended human body articulations. TS 15066 has introduced such an approach through (\ref{eq2}), which takes into account the variability and time-dependent nature of the position and speed of both the cobot and human co-worker \cite{b12}, as shown in Fig. \ref{fig2}.

\begin{equation}
    MSD(t_0)={S_H}+{S_R}+{S_S}+C+{Z_R}+{Z_D}
    \label{eq2}
\end{equation}

where, $Z_D$ represents the position of human co-worker, $Z_R$ denotes the position of the cobot's end-effector, and $S_S$ represents the stopping distance of the cobot, determined in accordance with ISO 10218.

   \begin{figure}[t]
      \centering
      \includegraphics[width=8.5cm]{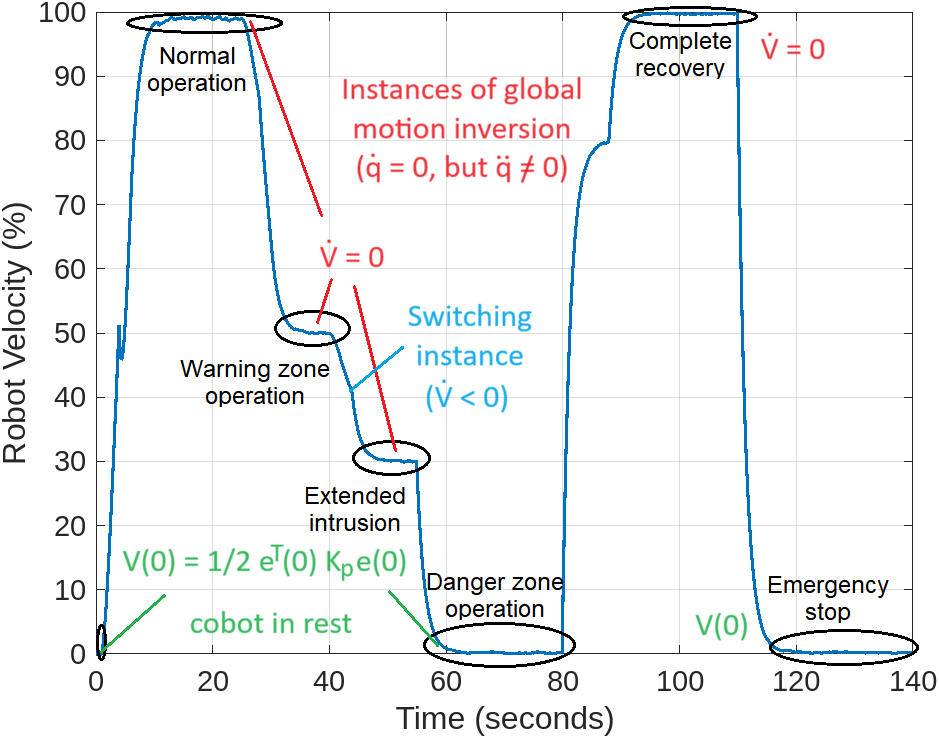}
      \caption{Velocity profile of cobot under collaborative conditions using proposed hierarchical controller formulation.}
      \label{figT}
   \end{figure}

\begin{equation*}
    S_S=V_R(t)\times(T_R+T_S)
\end{equation*}

$S_R$ represents the distance traveled by cobot before coming to rest after receiving stopping signal at the instant ($t_0$).

\begin{equation*}
    S_R=V_R(t)\times{T_R}
\end{equation*}

$S_H$ signifies the distance traveled by human co-worker before the cobot (i.e., both arm and gripper) comes to stop.

\begin{equation*}
    S_H=V_H(t)\times(T_R+T_S)
\end{equation*}
    
where, $V_H$ and $V_R$ represent speeds of human and cobot, respectively. Additionally, $T_R$ denotes the reaction time of cobot, and $T_S$ stands for response time of perception system.  

\subsection{Hierarchical Velocity Controller}

The proposed hierarchical velocity controller architecture comprises a primary and secondary speed regulation loops, as depicted in Fig. \ref{fig3}. The primary loop incorporates the proposed safety zone configurations and their intrusion status by utilizing distance computations from laser scanners (PSEN sc 5.5) to appropriately adjust the cobot's speed. Two lasers are strategically installed following the guidelines of EN 61496, as shown in Fig. \ref{figK} (a-c). The placement of laser scanners takes into consideration factors such as overlapping coverage, safe distance from the robot, mounting height, scanning range, and the field of view \cite{c8}. This configuration not only provides comprehensive coverage of collaborative workspace but also enables the detection of direction and inception of human operators approaching the cobot, further enhancing the safety and reliability of HRC setup. The output of these laser scanners is fed into installed safety PLC (PSS4000) to derive a common control signal for the cobot. The cobot controller features a real-time data exchange (RTDE) interface, implemented via URcaps \cite{b13}, which enables synchronization of external applications over a designated TCP/IP connection without compromising real-time performance. The primary control logic for regulating the cobot is governed by (\ref{eq5})

   \begin{figure*}[t]
      \centering
      \includegraphics[width=18cm]{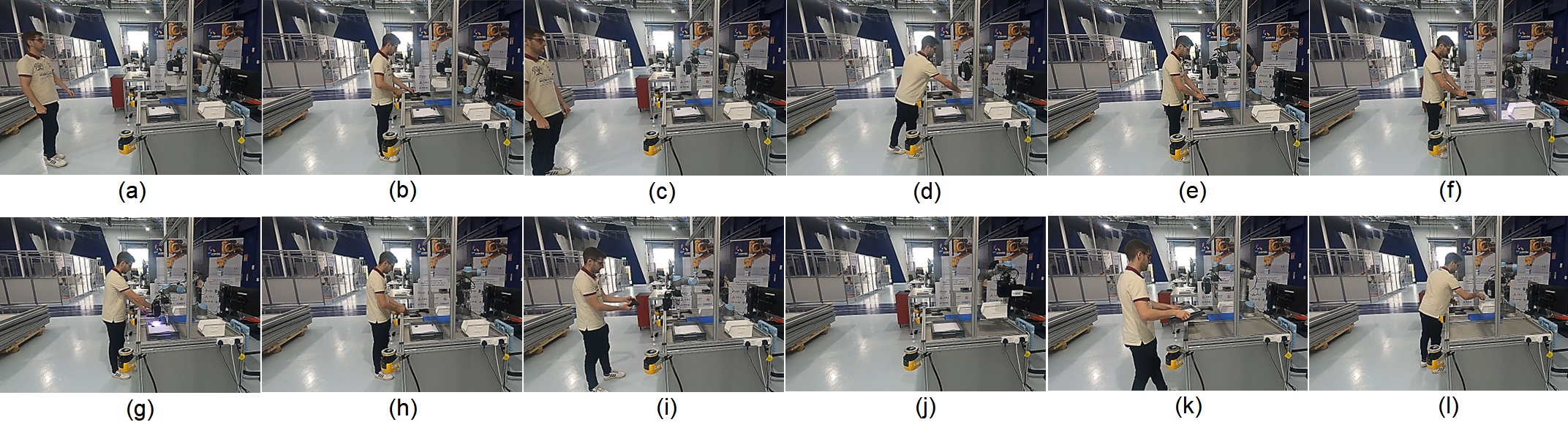}
      \caption{Experimental assessment of proposed solution within agile assembly settings. (a-d) represent human operator approaching shared workspace while cobot adapts it motion and configuration accordingly; (e-i) illustrate human operator performing assembly operation while cobot sorts parts into designated bins; (j-l) show how operator transit between subsequent zones for product batching while cobot arranges sorted bins into adjacent zone. }
      \label{figY}
   \end{figure*}

\begin{equation*}
    U(d)= \{
  \begin{array}{lr} 
      V_f & {Z^{min}_{norm}}\leq{d}\leq {Z^{max}_{norm}} \\
      V_c & {Z^{min}_{warn}}\leq{d}\leq {Z^{max}_{warn}} \\
      V_s & {d}\leq {Z^{max}_{dang}}
      \end{array}
\end{equation*}

\begin{equation}
    \Dot{e} = J^{\dagger}{\lambda}{V_l}+k_0(\pdv{(w(q))}{q})^{T}
    \label{eq5}
\end{equation}

\begin{equation*}
    u_q(t) = K_Pe(t) + K_D\dot{e}(t) 
\end{equation*}

where $U$ represents the control signal, $V$ stands for the Cartesian velocity (i.e., full (f), collaborative (c), and stand-still (s)) computed using laser scanners (l) to determine the distance $(d)$. Additionally, $Z$ denotes the zone configuration for normal (white), warning (yellow), and danger zones (red). $\Dot{e}$ represents the joint velocity error signal, $J^{\dagger}$ is the pseudo-inverse of the analytical Jacobian, $\lambda$ is a positive definite control gain matrix, $k_0$ is a positive scalar value, and $w(q)$ is a secondary objective function to optimize the cobot's performance. In this specific case, it pertains to the energy efficiency of joint motors regulated through current flow. $u_q(t)$ is PD control law in joint space, $K_P$ and $K_D$ are positive definite control gains matrices (i.e., tuned using Ziegler-Nicolas method), $e(t)$ and $\dot{e}(t)$ are the time evolution of joint position and velocity errors respectively.         

The secondary loop of this system utilizes the industrial-ROS interface of cobot to leverage human skeleton tracking landmarks generated through depth sensing module. This approach enables a further reduction in cobot's velocity, enhancing the resilience and robustness of proposed HRC setup. To achieve this, an industrial-grade skeleton tracking algorithm \cite{c7} is modified to generate additional depth markers along with original positional landmarks about the tracked points on human body in 3D space. This algorithm effectively tracks 32 points at a rate of 30 $Hz$, as depicted in Fig. \ref{figK} (d-f). These points are continuously evaluated to determine the distance between the cobot's tool center point and the closest point on the human body. This information is then used to modulate cobot's velocity through an effort-based velocity controller. The control law that governs the implementation of secondary loop is defined by (\ref{eqt6})

\begin{equation}
    U(d)= \{
  \begin{array}{lr} 
      {k_s(d)}V_c & {d_i}\leq {Z^{min}_{warn}}
      \end{array}
    \label{eqt6}
\end{equation}

\begin{equation*}
    \Dot{q} = {J}^{\dagger}{V_d}+(I_n-J^{\dagger}J)\Dot{q}_0
\end{equation*}

where $k_s(d)$ represents the scaling factor, $d_i$ is the Euclidean distance of closest point, $\Dot{q}$ is the joint velocity vector (where $\Dot{q}_0$ is nominal joint velocity vector), $I_n$ denotes the identity matrix, and $(I_n-J^{\dagger}J)$ is the null space projector. It is imperative to emphasize that the selection and fine-tuning of the control gains, (i.e., $K_P$, $K_D$, and $k_s$), are executed such that the primary loop possesses a higher bandwidth, enabling it to stabilize and respond with greater rapidity. Furthermore, these gains are determined to ensure smooth transitions, precluding the induction of overshoots, oscillations, or instabilities.  

\section{Stability Analysis}

In order to guarantee the safety, performance, and reliability of the proposed hierarchical control architecture, its global asymptotic stability is assessed utilizing LaSalle's theorem. This theorem posits the existence of a positive definite Lyapunov candidate $V$, for which the gradient is negative semi-definite $\dot{V}\leq 0$. Additionally, it postulates the existence of a maximal invariant set, denoted as $M$, within which the gradient of the Lyapunov candidate approaches zero, that is, $\dot{V}=0$. This principle is governed by (\ref{lyapunov})

\begin{equation*}
    V = {\frac{1}{2}}{e^T}{K_P}{e}+ {\frac{1}{2}}{\dot{e}}^T{K_D}{\dot{e}}
\end{equation*}

\begin{equation}
   M \subseteq S = \{x \in \mathbb{R} : \dot{V}(x) = 0 \}
    \label{lyapunov}
\end{equation}

Where, $S$ is the set of states along which the system trajectories ($x$) asymptotically converges to equilibrium. The $M$ is said to be largest invariant set if $\dot{V}=0$ given $x(t): \forall {t}{\geq} t_0 $ for ascertaining global asymptotic stability.   

Fig \ref{figT} illustrates the cobot's velocity profile during preliminary collaborative part inspection task, shown in Fig \ref{figK}. To assess the global stability of the proposed controller, the computed Lyapunov candidate $V$ is monitored throughout this trajectory. In the initial phase, when the operator remains within the normal safety zone, the cobot attains its maximum speed (1 $m/sec$), with $V$ being both positive definite and radially bounded. It indicates a stable rise in velocity (i.e., where $V > 0$) as the cobot undertakes the high-precision scanning task. However, as the commissioned laser scanners detect the operator's proximity, the velocity is reduced to almost 50\%, triggered by the primary loop in action. At this point, the condition $\dot{V}{\leq} 0$ underscores the system's asymptotic stability. Notably, despite the velocity reaching zero, a non-zero acceleration with intermediate plateaus highlights the persistent potential energy, even in the absence of kinetic energy, as described by (\ref{lyapunov}). When the operator moves into the warning zone, the depth camera-driven secondary loop further reduces the velocity to 30\% in response to attempted zone violations, according to ISO 10218-2. Subsequently, as the operator enters the danger zone, the cobot adopts a stand-still mode. Throughout these transitions, the invariant set $M$ consistently result into $\dot{V}=0$ . As the operator retreats from the danger zone, the cobot's acceleration is revived, impacting both the kinematic and potential energy elements within $V$. Nevertheless, the condition $\dot{V}\leq0$ persist, a testament to the asymptotic stability of the hierarchical controller. In situations necessitating an emergency stop, the emphasis shifts towards a swift reduction in speed, prioritizing safety. Owing to the inherent inertial damping, the system's exponential decay in motor energy reinforces its stable operation. Hence, the proposed hierarchical controller demonstrates global asymptotic equilibrium throughout all the transitions of uniquely defined safety zones. 
    
\section{Results and Discussion}

\begin{figure*}[t]
      \centering
      \includegraphics[width=18cm]{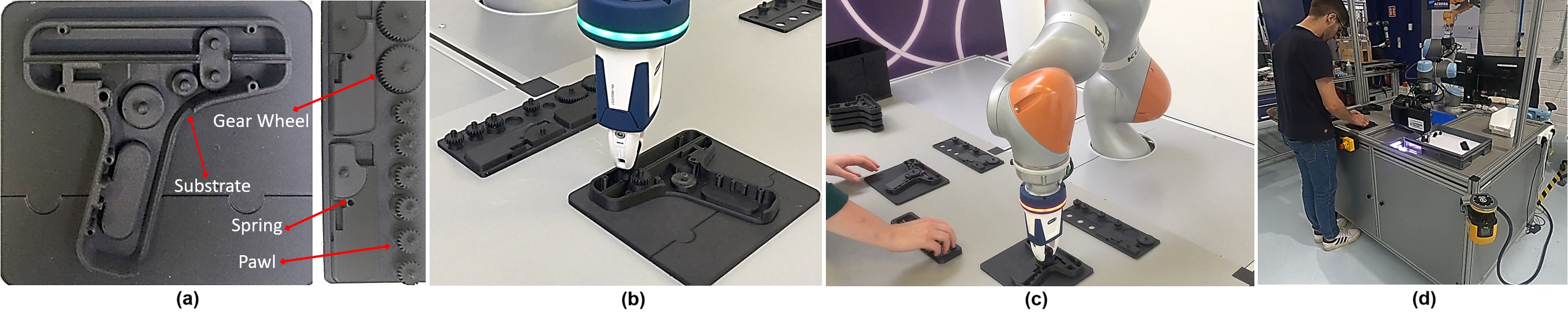}
      \caption{Experimental benchmarking of proposed HRC solution: (a) shows parts used in the stent gun assembly task; (b) represents autonomous cobot assembly without human intervention; (c) demonstrates traditional HRC setup with non-overlapping human-robot cooperation; (d) illustrates proposed HRC approach with simultaneous human-robot operations for efficient and resilient stent gun assembly.}
      \label{scenariofig}
   \end{figure*}

To validate the effectiveness and usability of the proposed solution, a collaborative scenario involving parts sorting and product assembly is executed using an industrial-grade collaborative robotic cell, as illustrated in Fig. \ref{figY}. The robotic cell is equipped with a cobot (UR5), a parallel gripper (OnRobot RG2), an eye-in-hand high-precision scanner (Zivid One+ S), two safety-rated laser scanners (Pilz sc M 5.5), eye-to-hand depth sensing (Intel D435i), and working medical parts. The task involves the assembly of a stent gun, typically used for angioplasty in cardiac patient treatment, and requires precise manipulation of gear wheels, pawls, springs, and plastic covers of various dimensions, as shown in Fig. \ref{scenariofig} (a). The cobot initiates the process by sorting these components into proper bins, passing them to the human-operator's instantaneous side, while simultaneously continuing to sort the remaining batch pieces to ensure continuity, uptime, and minimize deadlocks. Each unit is then assembled, tested, and packed by a human worker in an adjacent zone within the shared and dependable workspace. The safety monitoring zones are configured to cover approximately 1.5 meters in length and 900 millimeters in width of the collaborative workspace, in accordance with the guidelines of ISO 13855. The width of the zones is divided into two halves, creating right and left collateral quadrants, with the line of demarcation crossing the center of the base frame of the cobot. Furthermore, the cobot cell utilizes proposed hierarchical velocity controller that proactively regulates the cobot's speed based on the human movement in relation to cobot positions to avoid unintended contact and collisions, thereby realizing the SSM collaborative mode in accordance with ISO TS-10218. This hierarchical velocity controller, with primary and secondary loops, prioritizes laser scanner feedback using safety I/Os of main PLCs as the primary source of protection, while also facilitating monitoring of articulated motions of human co-worker using depth sensing, providing enhanced resilience and robustness to this collaborative setup.

\begin{figure}[t]
      \centering
      \includegraphics[width=8.5cm]{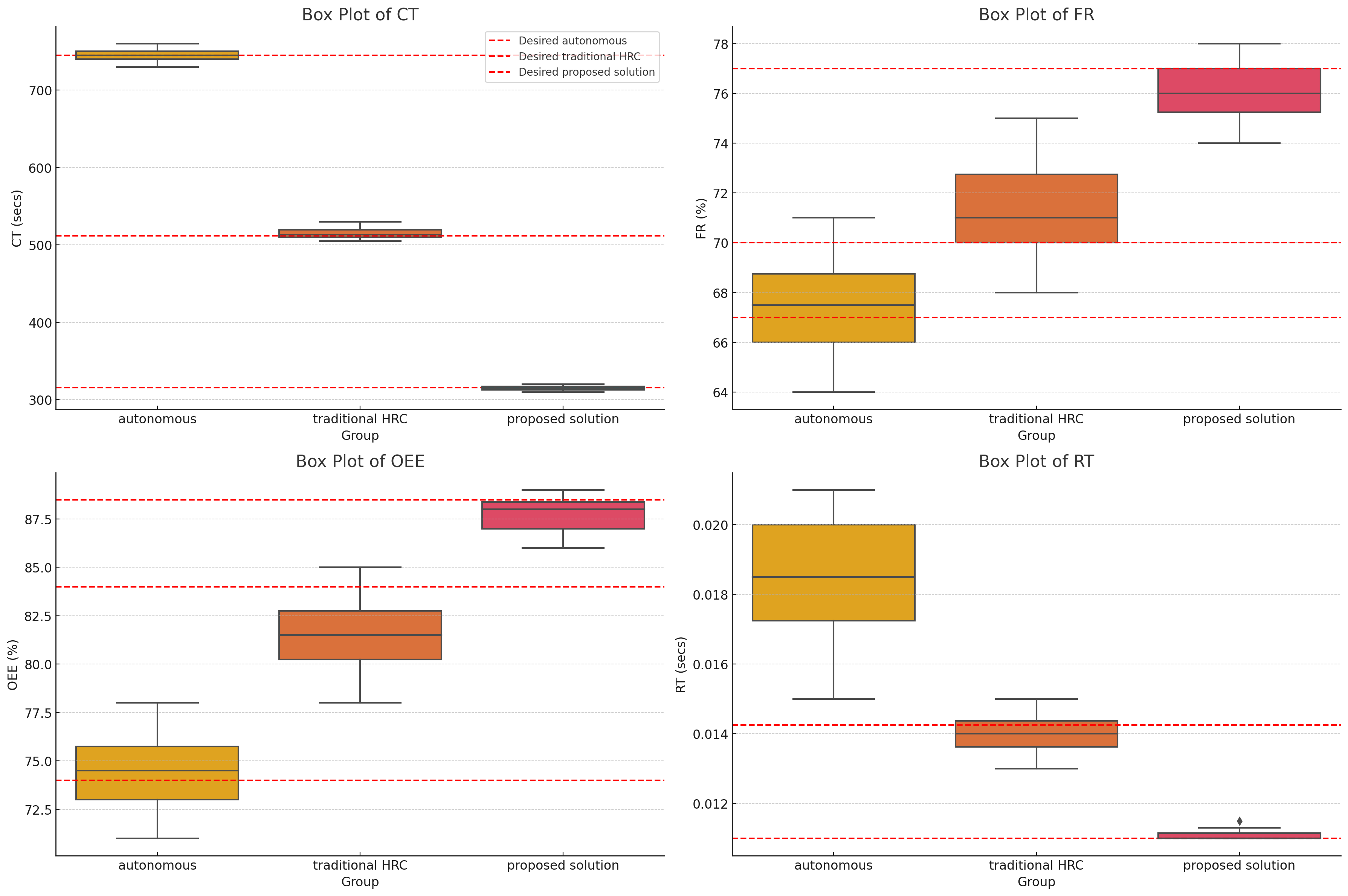}
      \caption{Performance benchmarking of proposed HRC solution against state of the art methods.}
      \label{figZZ}
   \end{figure}

To quantify the performance improvement resulting from the proposed solution, in terms of enhanced efficiency, productivity, resilience, and security, we recorded and reported four practical metrics: cycle time, overall equipment effectiveness, flexibility rate, and reaction delay, as illustrated in Fig \ref{figZZ}. The proposed methodology is benchmarked against two other state-of-the-arts, as shown in Fig. \ref{scenariofig}, to validate its merit and applicability into higher TRL agile industrial settings. The first method, termed 'autonomous,' describes an environment where the cobot assembles entire stent gun without human intervention. Due to intrinsic complexities of task and the cobot being the only agent, this led to extended assembly times. However, the second approach, denominated as "traditional HRC," employs standard PFL mode to safeguard the operator during concurrent activities. This method augments the safety, resilience, and dependability of the HRC arrangement; however, it inadvertently introduces disruptions and decelerates task execution due to the necessity of operating at collaborative speeds. In juxtaposition, the proposed solution, characterized by its novel zone configurations and amalgamated hierarchical velocity controller, manifests superior efficiency, resilience, and productivity, as substantiated by benchmarked data in Fig \ref{figZZ}.    

Our study indicates a significant reduction of 96 secs and 0.00325 secs in cycle time and reaction time, respectively, with the integration of the proposed solution compared to the traditional HRC solution. This improvement can be attributed to the fact that the cobot experiences fewer standstill conditions, and it continues its operation even when the human co-worker is assembling parts into the right quadrant for production, while the robot on the left quadrant performs sorting operation. This cooperative approach ensures desired collaboration and as well as separation, effectively overcoming undesired deadlocks and completing parallel tasks in a more efficient manner. Furthermore, the real-time running of the primary and secondary loops within the hierarchical velocity controller, synchronized with the main safety PLC, allows for instant switching and response, resulting in a reduced reaction time. The redundancy provided by depth sensing in addition to laser scanning also contributes to the system's enhanced resilience against unexpected events. Moreover, the integration of the proposed approach has led to a 7\% improvement in the flexibility rate and a 4.5\% improvement in the overall equipment effectiveness of the system respectively, compared to traditional settings. This is primarily due to the system's utilization of multi-sensing information to monitor human activities with respect to zone configuration and dynamically adapt to different configurations, thereby satisfying task constraints and optimizing system up-time for continuous production.

\section{Conclusions and Recommendations}

This study presents an innovative safety-oriented HRC strategy tailored for industrial-grade collaborative settings. The proposed framework employs a hierarchical velocity controller, which harnesses multi-modal perception derived from both 2D laser scanner feedback and 3D depth camera data. This integration facilitates precise regulation of the cobot’s speed, ensuring optimal system performance, stability, and workspace reliability without compromising the productivity and throughput of production line. The framework's efficacy has been validated under quasi-industrial conditions, benchmarked at TRL of 5, specifically for the collaborative assembly of medical devices. Furthermore, its asymptotic stability was rigorously analyzed using the Lasalle theorem. Empirical findings indicate that the introduced framework markedly diminishes task cycle times and controller response delays by 62\% and 29\%, respectively. Additionally, it augments system flexibility and overall equipment effectiveness by 7\% and 4.5\% in order, as compared to the traditional HRC approach.

For comprehensive evaluation of framework's utility, subsequent analyses will subject it to rigorous testing in extensive industrial manufacturing environments, as well as within digital twin settings. This will facilitate an assessment of its adaptability and transferability to commercial real-world scenarios, thereby benchmarking its scalability, reproducibility, and robustness across diverse and fluctuating conditions.\footnote{Performance of proposed setup have been meticulously evaluated across multiple distinct scenarios. For detailed results, please refer to; \url{https://drive.google.com/drive/folders/1GY5Q5B4iFTQYRg7Y6-I6aRWsNi76-6zo?usp=sharing}.}


\begin{thebibliography}{00}
\bibitem{b1} Patil, S., Vasu, V. and Srinadh, K.V.S., 2023. Advances and perspectives in collaborative robotics: a review of key technologies and emerging trends. Discover Mechanical Engineering, 2(1), p.13.
\bibitem{b2} https://ifr.org/downloads/press2018/2022$\_$WR$\_$extended$\_$version.pdf.
\bibitem{b3} https://www.interactanalysis.com/collaborative-robots-apr-2022/.
\bibitem{b4} Yang, C., Ganesh, G., Haddadin, S., Parusel, S., Albu-Schaeffer, A., \& Burdet, E. (2011). Human-like adaptation of force and impedance in stable and unstable interactions. IEEE transactions on robotics, 27(5), 918-930.
\bibitem{b6} Villani, V., Pini, F., Leali, F. and Secchi, C., 2018. Survey on human–robot collaboration in industrial settings: Safety, intuitive interfaces and applications. Mechatronics, 55, pp.248-266.
\bibitem{b7} Gopinath, V. and Johansen, K., 2019. Understanding situational and mode awareness for safe human-robot collaboration: case studies on assembly applications. Production Engineering, 13, pp.1-9.
\bibitem{b8} De Luca, A., Siciliano, B. and Zollo, L., 2005. PD control with on-line gravity compensation for robots with elastic joints: Theory and experiments. automatica, 41(10), pp.1809-1819.
\bibitem{b9} Matić, A., Valerjev, P. and Gomez-Marin, A., 2021. Hierarchical control of visually-guided movements in a 3D-printed robot arm. Frontiers in Neurorobotics, p.149.
\bibitem{b10}Ficuciello, F., Villani, L. and Siciliano, B., 2015. Variable impedance control of redundant manipulators for intuitive human–robot physical interaction. IEEE Transactions on Robotics, 31(4), pp.850-863.
\bibitem{c1} Sisbot, E.A. and Alami, R., 2012. A human-aware manipulation planner. IEEE Transactions on Robotics, 28(5), pp.1045-1057.
\bibitem{c1b} Long, P., Chevallereau, C., Chablat, D. and Girin, A., 2017. An industrial security system for human-robot coexistence. Industrial Robot: An International Journal, 45(2), pp.220-226.
\bibitem{c2} Gaz, C., Magrini, E. and De Luca, A., 2018. A model-based residual approach for human-robot collaboration during manual polishing operations. Mechatronics, 55, pp.234-247.
\bibitem{c3} Katyara, S., Deshpande, N., Ficuciello, F., Teng, T., Siciliano, B., Caldwell, D.G. and Chen, F., 2021. Formulating Intuitive Stack-of-Tasks using Visuo-Tactile Perception for Collaborative Human-Robot Fine Manipulation. arXiv preprint arXiv:2103.05676.
\bibitem{c4} Mohammadi Amin, F., Rezayati, M., van de Venn, H.W. and Karimpour, H., 2020. A mixed-perception approach for safe human–robot collaboration in industrial automation. Sensors, 20(21), p.6347.
\bibitem{c5} Wan, G., Dong, X., Dong, Q., He, Y. and Zeng, P., 2022. Design and implementation of agent-based robotic system for agile manufacturing: A case study of ARIAC 2021. Robotics and Computer-Integrated Manufacturing, 77, p.102349.
\bibitem{c6} D.	Karagiannis, P., Kousi, N., Michalos, G., Dimoulas, K., Mparis, K., Dimosthenopoulos, D., Tokçalar, Ö., Guasch, T., Gerio, G.P. and Makris, S., 2022. Adaptive speed and separation monitoring based on switching of safety zones for effective human robot collaboration. Robotics and Computer-Integrated Manufacturing, 77, p.102361.
\bibitem{b12}https://www.veobot.com/blog/2019/12/15/a-path-for-realistic-human-robot-collaborationpart-2.
\bibitem{c8} Michalos, G., Kousi, N., Karagiannis, P., Gkournelos, C., Dimoulas, K., Koukas, S., Mparis, K., Papavasileiou, A. and Makris, S., 2018. Seamless human robot collaborative assembly–An automotive case study. Mechatronics, 55, pp.194-211.
\bibitem{b13}Schmatz, F., Beuß, F., Sender, J. and Flügge, W., 2020. Use of human-robot collaboration to enhance process monitoring of mechanical joining. Procedia Manufacturing, 52, pp.272-276.
\bibitem{c7} Kim, J.W., Choi, J.Y., Ha, E.J. and Choi, J.H., 2023. Human pose estimation using mediapipe pose and optimization method based on a humanoid model. Applied Sciences, 13(4), p.2700.

\end{thebibliography}
\end{document}